\documentclass[conference,letterpaper]{IEEEtran}

\usepackage[utf8]{inputenc} 
\usepackage[T1]{fontenc}
\usepackage{url}
\usepackage{ifthen}
\usepackage[cmex10]{amsmath} 

\usepackage[utf8]{inputenc} 
\usepackage[T1]{fontenc}
\usepackage{url}             
\usepackage[sort, compress, space]{cite}            
\usepackage[cmex10]{amsmath}  
\usepackage{mleftright}       
\mleftright                   
\usepackage{graphicx}         
\usepackage{booktabs}         
\usepackage{algorithm, algorithmic}    
\usepackage[caption=false,font=footnotesize]{subfig}

\usepackage{amssymb, amsfonts} 
\usepackage{nicefrac}
\usepackage{mathtools}
\usepackage{bm, bbm}
\usepackage[scr=boondoxo,scrscaled=1.05]{mathalfa}
\usepackage{upgreek}

\usepackage{cleveref}

\usepackage{thm-restate}

\usepackage{dsfont}
\usepackage[shortlabels]{enumitem}
\usepackage{xcolor}
\usepackage{comment}
\usepackage{tikz}
\usepackage{tikzscale}
 
\usepackage{cuted}

\definecolor{orange}{rgb}{1,0.4,0.0}

\DeclarePairedDelimiterXPP{\KL}[2]{D_\textnormal{KL}}{(}{)}{}{%
#1\:\delimsize\|\:#2%
}
\DeclarePairedDelimiterXPP{\RD}[2]{D_{$\alpha$}}{(}{)}{}{%
#1\:\delimsize\|\:#2%
}

\DeclarePairedDelimiterXPP\Prob[1]{\mathbb{P}}{\lbrace}{\rbrace}{}{

#1}

\DeclarePairedDelimiterXPP{\lnorm}[2]{}{\lVert}{\rVert}{_{#2}}{#1}

\newcommand{\reg}{\ensuremath{\textnormal{reg}}}

\newcommand{\EREW}[1]{\ccR(#1)}

\newcommand{\IS}[2]{\widehat{\ccR}^{\mathrm{IS}}(#1,#2)}

\newcommand{\estimate}[2]{\widehat{\ccR}(#1,#2)}

\newcommand{\ISrew}[2]{\hat{r}^{\mathrm{IS}}(#1,#2)}

\newcommand{\CISrew}[3]{\hat{r}^{\mathrm{IS}}(#1,#2,#3)}

\newcommand{\expRew}[1]{\overline{r}(#1)}

\newcommand{\bE}{\ensuremath{\mathbb{E}}}

\newcommand{\bP}{\ensuremath{\mathbb{P}}}
\newcommand{\bQ}{\ensuremath{\mathbb{Q}}}
\newcommand{\bR}{\ensuremath{\mathbb{R}}}

\newcommand{\cA}{\ensuremath{\mathcal{A}}}
\newcommand{\cB}{\ensuremath{\mathcal{B}}}

\newcommand{\cE}{\ensuremath{\mathcal{E}}}

\newcommand{\cX}{\ensuremath{\mathcal{X}}}

\newcommand{\ccR}{\ensuremath{\mathscr{R}}}

\newtheorem{lemma}{Lemma}

\newtheorem{proposition}{Proposition}
\newtheorem{definition}{Definition}

\newtheorem{theorem}{Theorem}

\newtheorem{example}{Example}

\newcommand\xqed[1]{%
  \leavevmode\unskip\penalty9999 \hbox{}\nobreak\hfill
  \quad\hbox{#1}}

\newenvironment{proof}{\emph{Proof:}}{\xqed{$\blacksquare$}}

\newcommand{\argmax}{\operatornamewithlimits{argmax}}

\begin{document}
\title{Refined PAC-Bayes Bounds for Offline Bandits} 


\author{%
  \IEEEauthorblockN{Amaury Gouverneur, Tobias J. Oechtering, and Mikael Skoglund}
  \IEEEauthorblockA{Division of Information Science and Engineering (ISE)\\ 
                    KTH Royal Institute of Technology, Stockholm\\
                    \texttt{\{amauryg,oech,skoglund\}@kth.se}}
}

\maketitle


\begin{abstract}

In this paper, we present refined probabilistic bounds on empirical reward estimates for off-policy learning in bandit problems. We build on the PAC-Bayesian bounds from \cite{seldin2010pac, seldin_pac-bayesian_2011, seldin_pac-bayes-bernstein_nodate,seldin2012pac} and improve on their results using a new parameter optimization approach introduced by \cite{rodriguez-galvez_more_2024}. This technique is based on a discretization of the space of possible events to optimize the ``in probability'' parameter, $\lambda$. We provide two parameter-free PAC-Bayes bounds, one based on Hoeffding-Azuma's inequality and the other based on Bernstein's inequality. We prove that our bounds are \emph{almost optimal} as they recover the same rate as would be obtained by setting the ``in probability'' parameter after the realization of the data. \footnote{This work was supported by  the Wallenberg AI, Autonomous Systems and Software Program (WASP) funded by the Knut and Alice Wallenberg Foundation.}
\end{abstract}

\section{Introduction}
\label{sec:introduction}
In bandit problems, an agent  interacts sequentially for $T$ times with an unknown environment. At each time $t\in \{1,\ldots,T\}$,  the agent select an action $a_t \in \mathcal{A}$ according to a decision strategy $\pi$, a policy, expressed as a probability distribution over the possible actions. Based on the chosen action, the environment produces a reward. By learning from the observed action-reward pairs, the agent aims to identify actions that maximize the long-term cumulative reward. This decision-making framework has applications in diverse domains, including healthcare, finance, recommender systems, and telecommunications (see~\cite{bouneffouf_survey_2020, silva_multi-armed_2022} for a survey). \\

In the \emph{offline} bandit setting, the agent is provided with a dataset $h^t$ consisting of previously observed action-reward pairs, $\{a_n, r_n\}_{n=1}^t$. Based on this dataset, the agent must select a fixed policy that yields high rewards \emph{in expectation}. Since the reward distribution is unknown, the agent relies on an \emph{empirical estimate}, $\estimate{\pi}{h^t}$, to evaluate the expected reward associated with the policy $\pi$. A important question to guide the policy search is whether the empirical estimate $\estimate{\pi}{h^t}$ is close to the true expected reward $\EREW{\pi}$.\\

\emph{Probably approximately correct} (PAC) theory provides answers to this question by offering ``in probability'' guarantees on the difference between the empirical estimate and the true expectation. Traditionally, these bounds have been developed for supervised learning problems and depend on the complexity of the hypothesis space, which can be measured using quantities like the Vapnik–Chervonenkis dimension or Rademacher complexity, as discussed in \cite{shalev2014understanding}. PAC-Bayesian bounds generalize PAC guarantees and account for the dependence of the learned hypothesis on the observed dataset~\cite{shawe1996framework,mcallester1998some,mcallester1999pac,mcallester2003pac,catoni2007pac}. This dependence is typically measured using the relative entropy between a prior hypothesis and the learned hypothesis. \\

Initially focused on supervised learning problems where the dataset consists of independent and identically distributed samples, the PAC-Bayes framework was later extended to bandit problems~\cite{seldin2010pac}. This extension leverage martingale properties to address the dependencies inherent in sequential decision problems. The resulting bounds have found practical applications, as they motivated the introduction of relative entropy regularization methods. The idea is to search for a policy $\pi$ that maximizes the expected reward estimate while imposing a penalty on the relative entropy between $\pi$ and a prior policy $\mu$. Examples of such methods include Relative Entropy Policy Search \cite{peters_relative_2010}, Trust Region Policy Optimization \cite{schulman_trust_2017}, and Proximal Policy Optimization \cite{schulman_proximal_2017}.\\

One problematic of the PAC-Bayes bounds derived in \cite{seldin2010pac} and follow-up works \cite{seldin_pac-bayes-bernstein_nodate,seldin_pac-bayesian_2012,seldin_pac-bayesian_2011}  is that they depend on a parameter $\lambda>0$ that must be set before observing the data and cannot be optimized based on the data. \cite{seldin_pac-bayesian_2012} attempts to circumvent this difficulty deriving a bound that holds simultaneously for a grid of parameters, but their approach fells short to obtain a parameter-free bound with optimal rate.\\

In this work, we improve on these previous results and derive parameter-free PAC-Bayes bounds, achieving the optimal rate. We apply an optimization technique for the ``in probability'' parameter $\lambda$ that works by discretizing the \emph{space of possible events} for the bounds and optimizing the parameter $\lambda$ \emph{conditioned on the event} before applying a union bound~\cite{rodriguez-galvez_more_2024}. We provide two optimized PAC-Bayes bounds: one based on the Hoeffding-Azuma inequality and one on the Bernstein inequality. \\

The rest of the paper is organized as follows: 
\begin{itemize}
    \item Section~\ref{sec:preliminaries} introduces the notation, presents the multi-armed bandit problem and contextual bandit problem, and discusses the online and offline learning settings.
    \item Section~\ref{sec:pac-bayes} explains the importance sampling estimate before proving different PAC-Bayes bounds.
    \item Finally, our conclusion is presented in Section~\ref{sec:conclusion}.
\end{itemize}

\newpage
\section{Preliminaries}
\label{sec:preliminaries}
\subsection{General Notation}
\label{subsec:notation}

Random variables $X$ are written in capital letters, their realizations $x$ in lowercase letters, their outcome space in calligraphic letters $\cX$, and their distribution is written as $\bP_X$. The density of a random variable $X$ with respect to a measure $\mu$ is written as $f_X \coloneqq \frac{d\bP_X}{d\mu}$. When two (or more) random variables $X, Y$ are considered, the conditional distribution of $Y$ given $X$ is written as $\bP_{Y|X}$, and the notation is abused to write their joint distribution as $\bP_X \bP_{Y|X}$.
\\

We use the underscore notation $X_t$ to represent a random variable at time $t=1,\ldots, T$ and the exponent notation $X^t$ to denote a sequence of random variables $X^t \equiv (X_1,\ldots,X_t)$ for $t=2,\ldots,T$. For consistency we let $X^1 \equiv X_1$.\\

We use the notation $d\bP/d\bQ$ for the Radon-Nikodym derivative. The relative entropy between two probability distributions $\bP$ and $\bQ$ is defined as $\KL{\bP}{\bQ} := \int \log \big( \frac{d\bP}{d\bQ} \big) d\bP$ if $\bP$ is absolutely continuous with respect to $\bQ$ and $\KL{\bP}{\bQ} \to \infty$ otherwise. 

\subsection{Multi-Armed Bandits}
\label{subsec:multi-armed_bandits}

A \emph{multi-armed bandit} is a sequential decision problem where, at each time step $t \in \{1, \dots, T\}$, an agent chooses an action $A_t \in \mathcal{A}$ according to a decision policy $\pi_t \in \Pi$, expressed as a probability distribution over the possible actions; that is $\pi_t(a)$ expresses the probability that agent takes action $a$ at time $t$. The agent then receives a positive random reward $R_t \in [0,1]$ distributed according to some fixed but unknown reward distribution $\mathbb{P}_{R_t | A_t}$. As the reward distribution depends on the chosen action, it can be written as $R_t = R(A_t)$ for some random function $R$. We use the notation $\overline{r}(A_t) \coloneqq \bE[R(A_t)]$ to denote the expected reward associated with action $A_t$. The data is collected in a history $H^{t} = H^{t-1} \cup \{A_t, R_t\}$, where $H^{t-1}$ contains all action-reward pairs observed prior to $t$. \\

We introduce the notation $\EREW{\pi}$ to denote the expected reward obtained by taking actions according to a policy $\pi$, with the expectation taken both over the randomness of the action selection and the reward distribution: 
\begin{align}
    \EREW{\pi} \coloneqq \bE_{A \sim \pi(\cdot)} [\overline{r}(A)].
\end{align}

\begin{example}[Drug dosage]
\label{example:drug_trial}
In clinical trials for a new drug, selecting the dosage is critical. Each action $A_t \in \mathcal{A}$ represents a different dosage level, and the reward $R_t$ corresponds to the patient's response to the dosage, such as an improvement in health metrics or the absence of adverse side effects. 
\end{example}

\subsection{Contextual Bandits}
\label{subsec:contextual_bandits}

In a \emph{contextual bandit} problem, at each time step $t \in \{1,\ldots,T\}$, the agent first observes a context $X_t \in \cX$ sampled from a fixed but unknown distribution $\bP_X$. Based on the context, the agent selects an action $A_t \in \cA$ according to a decision policy. Here, the decision policy,  $\pi_t \in \Pi$, represents a probability distribution over the action given the observed context. That is, $\pi_t(a,x)$ gives the probability that the agent takes action $a$ given that he observed context $x$. The agent then receives a reward $R_t \in \bR$ from a fixed but unknown distribution that depends on the context and the action taken, $\mathbb{P}_{R_t | A_t, X_t}$. The reward can be written $R_t = R(A_t, X_t)$ for some random function $R$. We use the notation $\overline{r}(A_t,X_t) \coloneqq \bE[R(A_t,X_t)]$ to denote the expected reward for action $A_t$ and context $X_t$. The data is collected in a history $H^{t} = H^{t-1} \cup H_{t}$, where $H_{t} = \{ A_t, X_t, R_t \}$. Accordingly, we introduce the notation $\EREW{\pi}$ to denote the expected reward obtained by taking actions according to a policy $\pi$: 
\begin{align}
    \EREW{\pi} \coloneqq \bE_{X \sim \bP_X, A \sim \pi(\cdot, X) } [\overline{r}(A,X)].
\end{align}

\begin{example}[Online advertising]
\label{example:online}
In online advertising, advertisers select which ad to show to a user based on user data or context $X_t \in \mathcal{X}$, such as demographics, browsing history, or location. Each action $A_t \in \mathcal{A}$ corresponds to a different advertisement, and the reward $R_t$ represents an outcome, such as whether the user clicks on the ad or the revenue generated.
\end{example}

\subsection{Online and Offline Settings}
\label{subsec:online_offline}

In the \emph{online} setting, the goal of the agent is to find a \emph{sequence} of policies, $\pi_{1:T} =\{\pi_1,\pi_2, \ldots, \pi_T\}$ that maximizes the expected \emph{cumulative reward}, $\sum_{t=1}^{T} \EREW{\pi_t}$, or equivalently that minimizes the \emph{cumulative regret}, defined as the expected difference between the best cumulative reward and the cumulative reward obtained following the policies: 
\begin{align*}
    \reg(\pi_{1:T}) \coloneqq \sum_{t=1}^{T} \EREW{\pi^\star} - \EREW{\pi_t},
\end{align*}
where $\pi^\star \in \argmax_{\pi \in \Pi} \EREW{\pi}$ is an optimal policy. To ensure that such a policy exists, we make the technical assumption that the set of policies $\Pi$ is compact.  
In this setting, each policy $\pi_t$ is chosen based on the observed data $h^t$, and the agent faces the \emph{exploration-exploitation} dilemma: should it try less explored actions to learn more about the reward distribution or exploit the best-performing ones based on the information gathered. \\

In the \emph{offline}  setting, also known as batch learning \cite{swaminathan_counterfactual_2015},  the agent is given a fixed set of logged data $h^{t} = \{a_n, r_n\}_{n=1}^{t}$, where each action $A_n$ was sampled from a known logging policy $\pi_n$, also known as behavior policies, and each reward $r_n$ obtained from the unknown reward distribution $\mathbb{P}_{R|A=a_n}$. Note that if the logging policies $\pi_n$ are the same for all $n \in \{1,\ldots, t\}$  then the collected data consists of i.i.d. samples. Given the dataset, $h^t$, the agent's goal in the offline setting is to select a single  \emph{target} policy $\pi \in \Pi$ that improves on the logged policies and will yield the highest expected reward.\\

The choice to perform online or offline learning typically depends on the nature of the application. One could argue that \Cref{example:drug_trial} is better suited for offline learning than \Cref{example:online}.

\section{PAC-Bayes Bounds}
\label{sec:pac-bayes}

In both the online and offline setting, maximizing directly the expected reward $\EREW{\pi}$ over the policies $\pi \in \Pi$ is not possible as the reward distribution $\bP_R$ is unknown. To guide the policy search of a policy, we have to use \emph{empirical estimate} of the expected reward $\EREW{\pi}$ based on the observed data $H^t$.\\

One commonly used estimate is the importance sampling (IS) estimate \cite{sutton_reinforcement_1998}. For the multi-armed bandit setting, the importance sampling estimate for the expected reward associated with action $a\in \cA$ is given by: 
\begin{equation}
    \ISrew{a}{h^t} \coloneqq \frac{1}{t} \sum_{n= 1}^{t} \frac{\mathds{1}_a\{a_n\}}{\pi_n(a_n)}r_n, 
\end{equation} 
where the indicator function $\mathds{1}_a\{a_n\}$ returns $1$ if $a_n = a$ and $0$  otherwise. We can verify that $\ISrew{a}{h^t}$ is an unbiased estimator of $\expRew{a}$ as taking the expectation of the dataset $H^t$ gives $\bE[\ISrew{a}{H^t}] = \expRew{a}$. 

For contextual bandits, let $n(x,h^t) = \sum_{n=1}^t \mathds{1}_{x}\{x_n\}$ denote the number of times that the context $x\in \cX$ appeared in the dataset $h^t$. The importance sampling estimate of the expected reward associated with action $a \in \cA$ and context $x \in h^t$ is given by
\begin{align}
    \CISrew{a}{x}{h^t} \coloneqq 
        \frac{1}{n(x,h^t)} \sum_{n= 1}^{t} \frac{\mathds{1}_{(a,x)}\{(a_n,x_n)\}}{\pi_n(a_n,x_n)}r_n.
\end{align} 

When using these empirical estimates, one important question is whether the policy estimate $\IS{\pi}{h^t} = \bE_{A\sim \pi}[\ISrew{A}{h^t}]$ is close to the actual policy value $\EREW{\pi}$.  PAC-Bayes theory provides answers to that question as it offers in probability guarantees on the difference between $\EREW{\pi}$ and $\IS{\pi}{h^t}$.

\subsection{Bandit Problems and Martingales}
\label{subsec:martingales}
The bandit setting introduces challenges not encountered in typical PAC-Bayesian learning settings. In bandit problems, the data $H^t$ is often neither independent nor identically distributed, as it depends on the sequential actions which, in turn, depend on the previously observed data. These challenges can be addressed by working with \emph{martingales}.

\vspace{0.5em}
\begin{definition}[Martingale]
A sequence of random variables $M^t$ is called a martingale with respect to $X^t$ if the following conditions hold for every $n\in\{1,\ldots,t\}$:
\begin{enumerate}
    \item $M_n$ is completely determined by $X_1, \dots, X_n$,
    \item $\bE[|M_n|] < \infty$,
    \item $
    \bE[M_n \mid X_1, \dots, X_{n-1}] = M_{n-1}$. 
\end{enumerate}
\label{def:martingale}
\end{definition}
\vspace{0.5em}
The sequence of consecutive differences in a martingale sequence $Z_n = M_n-M_{n-1}$ is referred to as a \emph{martingale difference sequence} and satisfies $\bE[Z_n \mid X_1, \dots, X_{n-1}] = 0$.\\

One reason to work with the importance sampling estimate is that it naturally gives rise to martingale sequences. For instance, for any action $a \in \cA$, the sequence of random variables
\begin{equation} M_n^{{IS}}(a) = n\left(\ISrew{a}{H^n} - \expRew{a}\right), \label{eq:martingal_example} \end{equation}
forms a martingale sequence with respect to $H^t$ and 
\begin{equation} Z_n^{{IS}}(a) = \frac{\mathds{1}_a\{A_n\}}{\pi_n(A_n)}R_n - \expRew{a}, \label{eq:martingale_diff_example} \end{equation}
forms its corresponding martingale difference sequence. A proof of that statement is provided in \cite[Lemma B.1]{flynn_pac-bayes_2023}.

\subsection{PAC-Bayes Bound for $\IS{\pi}{H^t}$}
\label{subsec:PACB_for_martingales}

Deriving PAC-Bayes bounds for martingales can be done with two main tools: Donsker-Varadhan's variational formula for relative entropy \cite{donsker1975asymptotic} and a martingale concentration inequality, such as Hoeffding-Azuma inequality \cite{hoeffding_probability_nodate,azuma_weighted_1967} or Bernstein inequality \cite[Proposition 2.10]{wainwright2019high}. We recall these classical results below. 
\vspace{0.5em}
\begin{lemma}[Donsker-Varadhan's variational formula]
\label{lemma:DV}
    For any measurable, bounded function $h: \cA \to \bR$ and any probability distribution on $\cA$, $\mu \in \Pi$, such that $\bE_{A\sim \mu}[e^{h(A)}] < \infty$, we have 
    \begin{align*}
        \log \bE_{A\sim \mu}[e^{h(A)}] = \sup_{\pi \in \Pi} \left( \bE_{A\sim \pi}[h(A)] - \KL{\pi}{\mu} \right).
    \end{align*}
\end{lemma}
\vspace{0.5em}
\begin{lemma}[Hoeffding-Azuma's inequality]
\label{lemma:heoff}
    Let $Z^t$ be a martingale difference sequence, such that for all $n\in \{1,\ldots,t\}$ $Z_n \in [a_n, b_n]$ and let $M_t = \sum_{n=1}^t Z_n$ be the corresponding martingale. Then, for any $\lambda \in \bR$, we have
    \begin{align*}
        \bE[e^{\lambda M_t}]\leq e^{\frac{\lambda^2 \sum_{n=1}^t (b_n-a_n)^2}{8}}.
    \end{align*}
\end{lemma}
\vspace{0.5em}
\begin{lemma}[Bernstein's inequality]
\label{lemma:bern}

Let $Z^t$ be a martingale difference sequence such that $|Z_n|\leq [a,b]$ for all $n\in \{1,\ldots,t\}$ with probability $1$. Let $M_t = \sum_{n=1}^t Z_n$ be a corresponding martingale and $V_t = \sum_{n=1}^t \mathbb{E}[Z_n^2 \mid Z_1, \dots, Z_{n-1}]$ be the cumulative variance of this martingale. Then for any fixed $\lambda \in [0, \frac{1}{b-a}]$, we have
    \begin{align*}
        \bE[e^{\lambda M_t}]\leq \bE[e^{\lambda^2 V_t (e-2)}].
    \end{align*}
\end{lemma}
\vspace{0.5em}

The first PAC-Bayes bound we present was introduced in \cite{seldin_pac-bayesian_2011}, and as in \cite[Theorem 5]{seldin2012pac}, we provide an alternative proof based on Donsker-Varadhan variational formal and Hoeffding-Azuma's inequality.
\vspace{0.5em}
\begin{proposition}[Hoeffding PAC-Bayes bound for $\hat{r}^{\mathrm{IS}}$]
\label{prop:PACB-Hoeff-rIS}
    Let $\mu \in \Pi$ be any prior policy independent of $H^t$ and let $\varepsilon$ be a uniform lower bound on $\{\pi_n(a)\}_{n=1}^t$ for all $a \in \cA$ (resp. on $\{\pi_n(a,x)\}_{n=1}^t$ in the contextual bandits setting) . Then, for any $\lambda >0$, and any $\beta \in (0,1)$, with probability no smaller than $1-\beta$ (over the sampling of $H^t$)
    \begin{align*}
        |\EREW{\pi}-\IS{\pi}{H^t}| \leq \frac{\lambda}{8 t \varepsilon^2} + \frac{\KL{\pi}{\mu}+\ln \frac{2}{\beta}}{\lambda},
    \end{align*}
    holds simultaneously for all policy $\pi \in \Pi$ depending on $H^t$.   
\end{proposition}
\vspace{0.5em}
\begin{proof}
     We start by fixing $a\in \cA$ and observe that the elements of the martingale difference sequence $\{Z_n^{{IS}}(a)\}_{n=1}^t$, defined in \cref{eq:martingale_diff_example}, are in the interval $[-\expRew{a}, \frac{1}{\varepsilon}-\expRew{a}]$. Applying \Cref{lemma:heoff} on the corresponding martingale sequence $\{n\left(\ISrew{a}{H^n} - \expRew{a}\right)\}_{n=1}^t$,  for any $l>0$, we have that
     \begin{align*}
         \bE_{H^t}[e^{l t\left(\ISrew{a}{H^t} - \expRew{a}\right)}]\leq e^{\frac{l^2 t}{8 \varepsilon^2}}.
     \end{align*}
     We set $\lambda = l/t$ and integrate over $a\in \cA$ with respect to the distribution $\mu$, which gives
     \begin{align*}
         \bE_{A\sim \mu}\bE_{H^t}[e^{\lambda\left(\ISrew{A}{H^t} - \expRew{A}\right)}]\leq e^{\frac{\lambda^2}{8 t \varepsilon^2}}.
     \end{align*}
     Applying first Fubini’s theorem and then Donsker-Varadhan variational formula for $h(a)=\lambda\left(\ISrew{a}{h^t} - \expRew{a}\right)$, we get
     \begin{align*}
        \bE_{H^t}\left[e^{\sup_{\pi \in \Pi} \bE_{A \sim \pi}[\lambda\left(\ISrew{A}{H^t} - \expRew{A}\right)] - \KL{\pi}{\mu}  -\frac{\lambda^2}{8 t \varepsilon^2}}\right] \leq 1.
     \end{align*}
     The end of the proof uses Chernoff's bound. For any $\alpha>0$, with probability smaller than $e^{-\alpha}$ over $H^t$, we have  
     \begin{align*}
         \sup_{\pi \in \Pi} \lambda\left(\IS{\pi}{H^t}- \EREW{\pi}\right)-\KL{\pi}{\mu}-\frac{\lambda^2}{8 t \varepsilon^2} > \alpha .
     \end{align*}
     As the statement holds for the supremum over $\pi \in \Pi$, it then holds for any $\pi \in \Pi$, even those depending on the observed data $H^t$. We set $\beta/2 = e^{-\alpha}$, that is $\alpha = \ln(2/\beta)$, and take the complement. Rearranging terms, we get that with probability no smaller than $1-\beta/2$, for all $\pi \in \Pi$ simultaneously, we have  
     \begin{align*}
         \IS{\pi}{H^t}- \EREW{\pi} \leq \frac{\KL{\pi}{\mu}+ \ln(2/\beta) }{\lambda}+\frac{\lambda}{8 t \varepsilon^2} .
     \end{align*}
     Using the same argument to $\{n\left(\expRew{a}  - \ISrew{a}{H^n}\right)\}_{n=1}^t$ and a union bound over the two results concludes the proof.
\end{proof}\\

To derive a tighter bound in $\varepsilon$, one can use Bernstein's inequality instead of Hoeffding-Azuma's inequality and obtain a \emph{Bernstein PAC-Bayes bound} \cite[Theorem 7]{seldin2012pac}. 
\vspace{0.5em}
\begin{proposition}[Bernstein PAC-Bayes bound for $\hat{r}^{\mathrm{IS}}$]
\label{prop:PACB-Bern-rIS}
    Let $\mu \in \Pi$ be any prior policy independent of $H^t$ and let $\varepsilon$ be a uniform lower bound on $\{\pi_n(a)\}_{n=1}^t$ for all $a \in \cA$ (resp. on $\{\pi_n(a,x)\}_{n=1}^t$ in the contextual bandits setting) . Then, for any $\lambda \in (0,1)$, and any $\beta \in (0,1)$, with probability no smaller than $1-\beta$ (over the sampling of $H^t$)
    \begin{align*}
        |\EREW{\pi}-\IS{\pi}{H^t}| \leq \frac{\lambda 2 (e-2)}{t \varepsilon} + \frac{\KL{\pi}{\mu}+\ln \frac{2}{\beta}}{\lambda},
    \end{align*}
    holds simultaneously for all policy $\pi \in \Pi$ depending on $H^t$.   
\end{proposition}
\vspace{0.5em}
\begin{proof}
    Let $a\in \cA$ be a fixed action and let $Z^\mathrm{IS}_n(a)$ be as defined in \cref{eq:martingale_diff_example}. We define  $V^\mathrm{IS}_t(a) \coloneqq \sum_{n=1}^t \bE[(Z^\mathrm{IS}_n(a))^2|Z^\mathrm{IS}_1(a),\ldots,Z^\mathrm{IS}_{n-1}(a)]$ the variance of the martingale $M_t^\mathrm{IS}(a)$ defined in \cref{eq:martingal_example}. The proof then follows similarly as the proof of \Cref{prop:PACB-Hoeff-rIS} using Bernstein's inequality instead of Hoeffding-Azuma's inequality. The final result is obtained using \cite[Lemma 2]{seldin_pac-bayes-bernstein_nodate} to get the upper bound $\bE_{A\sim \pi}[V^\mathrm{IS}_t(A)] \leq \frac{2 t}{\varepsilon}$ on the expected variance of the martingale.
\end{proof}

\subsection{Parameter-free PAC-Bayes bound for $\IS{\pi}{H^t}$}

Under their current forms, the bounds in \Cref{prop:PACB-Hoeff-rIS} and \Cref{prop:PACB-Bern-rIS} are not readily usable; we still need to choose a value of $\lambda>0$. In \Cref{prop:PACB-Hoeff-rIS}, if we could select the parameter $\lambda$ after observing $H^t$, we would choose the parameter that minimizes the bound that is we would use $\lambda = 2\varepsilon\sqrt{2t(\KL{\pi}{\mu} + \ln(2/\beta))}$. For this value, the bound in \Cref{prop:PACB-Hoeff-rIS} becomes $$| \IS{\pi}{H^t}- \EREW{\pi} | \leq \frac{1}{\varepsilon}\sqrt{\frac{\KL{\pi}{\mu} + \ln(2/\beta)}{2t}}.$$

Similarly, if we could set $\lambda = \sqrt{ \frac{t \varepsilon(\KL{\pi}{\mu} + \ln(2/\beta))}{2 (e-2)}}$ in \Cref{prop:PACB-Bern-rIS}, we would optimize the right-hand side and obtain a bound of $$| \IS{\pi}{H^t}- \EREW{\pi} | \leq \sqrt{\frac{8 (e-2)(\KL{\pi}{\mu} + \ln(2/\beta))}{t\varepsilon}}.$$ 

However, this is not possible since the parameter needs to be selected \emph{before} the draw of $H^t$ and can, therefore, not depend on the realization of this data \cite[Remark 14]{banerjee2021information}. \\

One idea to circumvent this difficulty is to consider a grid of values $\{\lambda_i\}_{i=1}^K$ of $\lambda$ such that the corresponding bound holds and with probability $\{1- \beta\}_{i=1}^K$. Applying a union bound argument, one gets a bound holds for all $\{\lambda_i\}_{i=1}^K$ simultaneously with probability at least $1-\sum_{i=1}^K \beta_i$. This technique is used in \cite{seldin_pac-bayesian_2012} but fails to attain a rate as tight as the one associated with the optimal value of $\lambda$. Our next results improve on the previous approach and provide \emph{almost optimal} bounds achieving the same rate as the optimal bound. 
\vspace{0.5em}
\begin{theorem}[Optimized Hoeffding PAC-Bayes bound]
\label{thm:optimized_hoeffding}
    Let $\mu \in \Pi$ be any prior policy independent of $H^t$ and let $\varepsilon$ be a uniform lower bound on $\{\pi_n(a)\}_{n=1}^t$ for all $a \in \cA$. Then, for any $\beta \in (0,1)$, with probability no smaller than $1-\beta$ (over the sampling of $H^t$), we have that
    \begin{align*}
        |\EREW{\pi}-\IS{\pi}{H^t}| \leq \frac{1}{\varepsilon}\sqrt{\frac{\KL{\pi}{\mu} + \ln \frac{4 \uppi}{3 \beta}}{t}},
    \end{align*}
    holds simultaneously for all policy $\pi \in \Pi$ depending on $H^t$.
\end{theorem}
\vspace{0.5em}
\begin{proof}
    Our proof is based on the technique introduced by \cite{rodriguez-galvez_more_2024} and refined in \cite{rodriguez-galvez_information-theoretic_2024}. We start with the parametric bound from  \Cref{prop:PACB-Hoeff-rIS}. At a high level, the idea is to form a grid over the events’ space (that is, the possible values for $\KL{\pi}{\mu}$) and find the best parameter conditioned on each event in that grid.
    To do so, we define the event $\cE_1 = \{\KL{\pi}{\mu} \leq 1\}$ and for all $k \in \{2,\ldots\}$, we set the event $\cE_k = \{k-1 < \KL{\pi}{\mu} \leq k\}$.
    Conditioned on the event $\cE_k$, there are two parameters that we can tune, $\lambda_k >0$ and $\beta_k \in [0,1]$. We then define the event $\cB_{\lambda_k, \beta_k}$ as the set $\{H^t : |\EREW{\pi}-\IS{\pi}{H^t}| > \frac{\lambda_k}{8 t \varepsilon^2} + \frac{\KL{\pi}{\mu}+\ln \frac{2}{\beta_k}}{\lambda_k}\}$.
    Then given the event $\cE_k$, with probability no more than $\bP[\cB_{\lambda_k, \beta_k}|\cE_k]$, there exists a dataset $H^t$ such that 
    $$|\EREW{\pi}-\IS{\pi}{H^t}| > \frac{\lambda_k}{8 t \varepsilon^2} + \frac{k+\ln \frac{2}{\beta_k}}{\lambda_k},$$
    where we used the fact that $\KL{\pi}{\mu}\leq k$ given $\cE_k$.
    As the right hand-side does not depend on the data $H^t$ anymore, we can optimize over $\lambda_k$. We first set $\beta_k = \frac{6 \beta}{\uppi k^2}$ (note that $\uppi$ denotes here the mathematical constant pi and not the policy as on the left hand-side of the inequality) and then set the optimal value $\lambda_k = \sqrt{8t \varepsilon (k+ \ln \frac{2 \uppi k^2}{6 \beta})}$. We get that with probability no more than $\bP[\cB_{\lambda_k, \beta_k}|\cE_k]$, 
    $$|\EREW{\pi}-\IS{\pi}{H^t}| > \frac{1}{\varepsilon}\sqrt{\frac{k+ \ln \frac{\uppi k^2}{3 \beta}}{2 t} }.$$
     The square root is a non-decreasing function, and given $\cE_k$, we have $k<\KL{\pi}{\mu}+1$. It therefore comes, given $\cE_k$, with probability no more than $\bP[\cB_{\lambda_k, \beta_k}|\cE_k]$, 
    $$|\EREW{\pi}-\IS{\pi}{H^t}| > \sqrt{\frac{\KL{\pi}{\mu} \smash{+} \ln \frac{e \uppi (\KL{\pi}{\mu}+1)^2}{3 \beta}}{2 t \varepsilon^2} }.$$
    Since $x +  \ln(\frac{e\uppi(1+x)^2}{3\beta})$ is a non-decreasing, concave, continuous function for all $x > 0$, it can be upper bounded by its envelope. That is $x +  \ln(\frac{e\uppi(1+x)^2}{3\beta}) \leq \inf_{a>0}\Big\{\frac{a+3}{a+1}x+\ln(\frac{e\uppi(a+1)^2}{3})-\frac{2a}{a+1}\Big\}$. Setting $a=1$, we get $x +  \ln(\frac{e\uppi(1+x)^2}{3\beta})  \leq 2x + \ln(\frac{4\uppi}{3\beta})$. Using this bound, we get that given $\cE_k$, with probability no more than $\bP[\cB_{\lambda_k, \beta_k}|\cE_k]$, 
    \begin{equation}
    \label{eq:event}
        |\EREW{\pi}-\IS{\pi}{H^t}| > \frac{1}{\varepsilon}\sqrt{\frac{\KL{\pi}{\mu}+\ln \frac{4 \uppi}{3 \beta}}{t} }.
    \end{equation}
    We now define $\cB'$ as the event described in~\cref{eq:event}. We have $\bP[\cB'|\cE_k]\bP[\cE_k] \leq \bP[\cB_k|\cE_k]\bP[\cE_k] \leq \bP[\cB_k] \leq \beta_k = \frac{6 \beta}{\uppi k^2}$. Therefore the probability of $\cB'$ is bounded as $\bP[\cB'] = \sum_{k=1}^\infty \bP[\cB'|\cE_k]\bP[\cE_k] \leq \sum_{k=1}^\infty \frac{6 \beta}{\uppi k^2} = \beta$. We then have $\bP[\cB'^C]=1-\bP[\cB'] \geq 1-\beta$ which concludes our proof. 
    
\end{proof}
\vspace{0.5em}
\begin{theorem}[Optimized Bernstein PAC-Bayes bound]
\label{thm:optimized_bernstein}
    Let $\mu \in \Pi$ be any prior policy independent of $H^t$ and let $\varepsilon$ be a uniform lower bound on $\{\pi_n(a)\}_{n=1}^t$ for all $a \in \cA$. Then for any $\beta \in (0,1)$, with probability no smaller than $1-\beta$ (over the sampling of $H^t$), we have that
    \begin{align*}
        |\EREW{\pi}-\IS{\pi}{H^t}| \leq 2\sqrt{\frac{(e-2)(\KL{\pi}{\mu}+\ln \frac{4 \uppi}{3 \beta})}{t\varepsilon} },
    \end{align*}
    holds simultaneously for all policy $\pi \in \Pi$ such that $\KL{\pi}{\mu}$ is smaller than $\frac{2(e-2)-t\varepsilon(2+\ln \frac{ 2 }{\beta})}{2t\varepsilon}$.
\end{theorem}
\vspace{0.5em}
\begin{proof}
    The proof follows by applying the same technique as for \Cref{thm:optimized_hoeffding}. As by assumption $\KL{\pi}{\mu} \leq \frac{2(e-2)-t\varepsilon(2+\ln \frac{ 2}{\beta})}{2t\varepsilon}$, we can define the event the events $\cE_k = \{k-1 < \KL{\pi}{\mu} \leq k\}$ only for $k\in\{1,\ldots,K\}$,  where $K = \left\lceil \frac{2(e-2)-t\varepsilon(2+\ln \frac{ \uppi }{3\beta})}{2t\varepsilon} \right\rceil $. Then we set the event $\cB_{\lambda_k, \beta_k}$ as the set $\{H^t : |\EREW{\pi}-\IS{\pi}{H^t}| > \frac{\lambda_k 2 (e-2)}{t \varepsilon} + \frac{\KL{\pi}{\mu}+\ln \frac{2}{\beta_k}}{\lambda_k}\}$. 
    Then given the event $\cE_k$, with probability no more than $\bP[\cB_{\lambda_k, \beta_k}|\cE_k]$, there exists a dataset $H^t$ such that 
    $$|\EREW{\pi}-\IS{\pi}{H^t}| > \frac{\lambda_k 2 (e-2)}{t \varepsilon} + \frac{k+\ln \frac{2}{\beta_k}}{\lambda_k},$$
    where we used the fact that $\KL{\pi}{\mu}\leq k$ given $\cE_k$. As before, we set $\beta_k = \frac{6 \beta}{\uppi k^2}$ and optimize over $\lambda_k$, that is we  set the optimal value $\lambda_k = \sqrt{\frac{t \varepsilon (k+ \ln \frac{2 \uppi k^2}{6 \beta})}{2(e-2)}}$. The value of $\lambda_k$ is the largest for $k=K$, and we can verify that $\lambda_K = \sqrt{\frac{t \varepsilon (K+ \ln \frac{2 \uppi K^2}{6 \beta})}{2(e-2)}} \leq \sqrt{\frac{t \varepsilon (2K+1 +\ln \frac{ \uppi }{3\beta})}{2(e-2)}} \leq 1$. The proof then concludes in a similar way as for \Cref{thm:optimized_hoeffding}.
\end{proof}

\section{Conclusion}
\label{sec:conclusion}

In this paper, we provide refined PAC-Bayes bounds for bandit problems, extending the analysis from~\cite{seldin_pac-bayesian_2011}. We propose two parameter-free bounds, one based on Hoeffding-Azuma inequality, the other on the Bernstein inequality, that achieve optimal rates through a new ``in probability'' parameter optimization technique. Future research will focus on leveraging these refined reward estimate bounds to derive new PAC-Bayes regret bounds.


\IEEEtriggeratref{21}

\bibliographystyle{IEEEtran}
\bibliography{references_isit,extra_references}

\begin{thebibliography}{10}
\providecommand{\url}[1]{#1}
\csname url@samestyle\endcsname
\providecommand{\newblock}{\relax}
\providecommand{\bibinfo}[2]{#2}
\providecommand{\BIBentrySTDinterwordspacing}{\spaceskip=0pt\relax}
\providecommand{\BIBentryALTinterwordstretchfactor}{4}
\providecommand{\BIBentryALTinterwordspacing}{\spaceskip=\fontdimen2\font plus
\BIBentryALTinterwordstretchfactor\fontdimen3\font minus \fontdimen4\font\relax}
\providecommand{\BIBforeignlanguage}[2]{{%
\expandafter\ifx\csname l@#1\endcsname\relax
\typeout{** WARNING: IEEEtran.bst: No hyphenation pattern has been}%
\typeout{** loaded for the language `#1'. Using the pattern for}%
\typeout{** the default language instead.}%
\else
\language=\csname l@#1\endcsname
\fi
#2}}
\providecommand{\BIBdecl}{\relax}
\BIBdecl

\bibitem{seldin2010pac}
Y.~Seldin and N.~Tishby, ``{PAC}-{B}ayesian analysis of co-clustering and beyond.'' \emph{Journal of Machine Learning Research}, vol.~11, no.~12, 2010.

\bibitem{seldin_pac-bayesian_2011}
\BIBentryALTinterwordspacing
Y.~Seldin, F.~Laviolette, J.~Shawe-Taylor, J.~Peters, and P.~Auer, ``\BIBforeignlanguage{en}{{PAC}-{Bayesian} {Analysis} of {Martingales} and {Multiarmed} {Bandits}},'' May 2011, arXiv:1105.2416 [cs]. [Online]. Available: \url{http://arxiv.org/abs/1105.2416}
\BIBentrySTDinterwordspacing

\bibitem{seldin_pac-bayes-bernstein_nodate}
Y.~Seldin, N.~Cesa-Bianchi, P.~Auer, F.~Laviolette, and J.~Shawe-Taylor, ``\BIBforeignlanguage{en}{{PAC}-{Bayes}-{Bernstein} {Inequality} for {Martingales} and its {Application} to {Multiarmed} {Bandits}}.''

\bibitem{seldin2012pac}
Y.~Seldin, F.~Laviolette, N.~Cesa-Bianchi, J.~Shawe-Taylor, and P.~Auer, ``{PAC}-{B}ayesian inequalities for martingales,'' \emph{IEEE Transactions on Information Theory}, vol.~58, no.~12, pp. 7086--7093, 2012.

\bibitem{rodriguez-galvez_more_2024}
\BIBentryALTinterwordspacing
B.~Rodríguez-Gálvez, R.~Thobaben, and M.~Skoglund, ``\BIBforeignlanguage{en}{More {PAC}-{Bayes} bounds: {From} bounded losses, to losses with general tail behaviors, to anytime validity},'' Jun. 2024, arXiv:2306.12214 [stat]. [Online]. Available: \url{http://arxiv.org/abs/2306.12214}
\BIBentrySTDinterwordspacing

\bibitem{bouneffouf_survey_2020}
\BIBentryALTinterwordspacing
D.~Bouneffouf, I.~Rish, and C.~Aggarwal, ``\BIBforeignlanguage{en}{Survey on {Applications} of {Multi}-{Armed} and {Contextual} {Bandits}},'' in \emph{\BIBforeignlanguage{en}{2020 {IEEE} {Congress} on {Evolutionary} {Computation} ({CEC})}}.\hskip 1em plus 0.5em minus 0.4em\relax Glasgow, United Kingdom: IEEE, Jul. 2020, pp. 1--8. [Online]. Available: \url{https://ieeexplore.ieee.org/document/9185782/}
\BIBentrySTDinterwordspacing

\bibitem{silva_multi-armed_2022}
\BIBentryALTinterwordspacing
N.~Silva, H.~Werneck, T.~Silva, A.~C. Pereira, and L.~Rocha, ``\BIBforeignlanguage{en}{Multi-{Armed} {Bandits} in {Recommendation} {Systems}: {A} survey of the state-of-the-art and future directions},'' \emph{\BIBforeignlanguage{en}{Expert Systems with Applications}}, vol. 197, p. 116669, Jul. 2022. [Online]. Available: \url{https://linkinghub.elsevier.com/retrieve/pii/S0957417422001543}
\BIBentrySTDinterwordspacing

\bibitem{shalev2014understanding}
S.~Shalev-Shwartz and S.~Ben-David, \emph{Understanding machine learning: {F}rom theory to algorithms}.\hskip 1em plus 0.5em minus 0.4em\relax Cambridge university press, 2014.

\bibitem{shawe1996framework}
J.~Shawe-Taylor, P.~L. Bartlett, R.~C. Williamson, and M.~Anthony, ``A framework for structural risk minimisation,'' in \emph{Conference on Computational learning theory (COLT)}, 1996, pp. 68--76.

\bibitem{mcallester1998some}
D.~A. McAllester, ``Some {PAC}-{B}ayesian theorems,'' in \emph{Conference on Computational learning theory (COLT)}, 1998, pp. 230--234.

\bibitem{mcallester1999pac}
------, ``{PAC}-{B}ayesian model averaging,'' in \emph{Conference on Computational learning theory (COLT)}, 1999, pp. 164--170.

\bibitem{mcallester2003pac}
------, ``{PAC}-{B}ayesian stochastic model selection,'' \emph{Machine Learning}, vol.~51, no.~1, pp. 5--21, 2003.

\bibitem{catoni2007pac}
O.~Catoni, ``{PAC}-{B}ayesian supervised classification: {T}he thermodynamics of statistical learning,'' \emph{IMS Lecture Notes Monograph Series}, vol.~56, p. 163pp, 2007.

\bibitem{peters_relative_2010}
\BIBentryALTinterwordspacing
J.~Peters, K.~Mulling, and Y.~Altun, ``\BIBforeignlanguage{en}{Relative {Entropy} {Policy} {Search}},'' \emph{\BIBforeignlanguage{en}{Proceedings of the AAAI Conference on Artificial Intelligence}}, vol.~24, no.~1, pp. 1607--1612, Jul. 2010. [Online]. Available: \url{https://ojs.aaai.org/index.php/AAAI/article/view/7727}
\BIBentrySTDinterwordspacing

\bibitem{schulman_trust_2017}
\BIBentryALTinterwordspacing
J.~Schulman, S.~Levine, P.~Moritz, M.~I. Jordan, and P.~Abbeel, ``\BIBforeignlanguage{en}{Trust {Region} {Policy} {Optimization}},'' Apr. 2017, arXiv:1502.05477 [cs]. [Online]. Available: \url{http://arxiv.org/abs/1502.05477}
\BIBentrySTDinterwordspacing

\bibitem{schulman_proximal_2017}
\BIBentryALTinterwordspacing
J.~Schulman, F.~Wolski, P.~Dhariwal, A.~Radford, and O.~Klimov, ``\BIBforeignlanguage{en}{Proximal {Policy} {Optimization} {Algorithms}},'' Aug. 2017, arXiv:1707.06347 [cs]. [Online]. Available: \url{http://arxiv.org/abs/1707.06347}
\BIBentrySTDinterwordspacing

\bibitem{seldin_pac-bayesian_2012}
\BIBentryALTinterwordspacing
Y.~Seldin, F.~Laviolette, N.~Cesa-Bianchi, J.~Shawe-Taylor, and P.~Auer, ``\BIBforeignlanguage{en}{{PAC}-{Bayesian} {Inequalities} for {Martingales}},'' Jul. 2012, arXiv:1110.6886 [cs]. [Online]. Available: \url{http://arxiv.org/abs/1110.6886}
\BIBentrySTDinterwordspacing

\bibitem{swaminathan_counterfactual_2015}
\BIBentryALTinterwordspacing
A.~Swaminathan and T.~Joachims, ``\BIBforeignlanguage{en}{Counterfactual {Risk} {Minimization}},'' in \emph{\BIBforeignlanguage{en}{Proceedings of the 24th {International} {Conference} on {World} {Wide} {Web}}}.\hskip 1em plus 0.5em minus 0.4em\relax Florence Italy: ACM, May 2015, pp. 939--941. [Online]. Available: \url{https://dl.acm.org/doi/10.1145/2740908.2742564}
\BIBentrySTDinterwordspacing

\bibitem{sutton_reinforcement_1998}
R.~S. Sutton and A.~G. Barto, \emph{\BIBforeignlanguage{en}{Reinforcement learning: an introduction}}, ser. Adaptive computation and machine learning.\hskip 1em plus 0.5em minus 0.4em\relax Cambridge, Mass: MIT Press, 1998.

\bibitem{flynn_pac-bayes_2023}
\BIBentryALTinterwordspacing
H.~Flynn, D.~Reeb, M.~Kandemir, and J.~Peters, ``\BIBforeignlanguage{en}{{PAC}-{Bayes} {Bounds} for {Bandit} {Problems}: {A} {Survey} and {Experimental} {Comparison}},'' \emph{\BIBforeignlanguage{en}{IEEE Transactions on Pattern Analysis and Machine Intelligence}}, vol.~45, no.~12, pp. 15\,308--15\,327, Dec. 2023, arXiv:2211.16110 [cs, stat]. [Online]. Available: \url{http://arxiv.org/abs/2211.16110}
\BIBentrySTDinterwordspacing

\bibitem{donsker1975asymptotic}
M.~D. Donsker and S.~S. Varadhan, ``Asymptotic evaluation of certain {M}arkov process expectations for large time, {I},'' \emph{Communications on Pure and Applied Mathematics}, vol.~28, no.~1, pp. 1--47, 1975.

\bibitem{hoeffding_probability_nodate}
W.~Hoeffding, ``\BIBforeignlanguage{en}{{PROBABILITY} {INEQUALITIES} {FOR} {SUMS} {OF} {BOUNDED} {RANDOM} {VARIABLESl}}.''

\bibitem{azuma_weighted_1967}
K.~Azuma, ``Weighted {Sums} of {Certain} {Dependent} {Random} {Variables},'' \emph{Tohoku Mathematical Journal, Second Series}, vol.~19, no.~3, pp. 357--367, 1967.

\bibitem{wainwright2019high}
M.~J. Wainwright, \emph{High-dimensional statistics: {A} non-asymptotic viewpoint}.\hskip 1em plus 0.5em minus 0.4em\relax Cambridge University Press, 2019, vol.~48.

\bibitem{banerjee2021information}
P.~K. Banerjee and G.~Mont{\'u}far, ``Information complexity and generalization bounds,'' in \emph{2021 IEEE International Symposium on Information Theory (ISIT)}.\hskip 1em plus 0.5em minus 0.4em\relax IEEE, 2021, pp. 676--681.

\bibitem{rodriguez-galvez_information-theoretic_2024}
\BIBentryALTinterwordspacing
B.~Rodríguez-Gálvez, R.~Thobaben, and M.~Skoglund, ``\BIBforeignlanguage{en}{An {Information}-{Theoretic} {Approach} to {Generalization} {Theory}},'' Aug. 2024, arXiv:2408.13275 [stat]. [Online]. Available: \url{http://arxiv.org/abs/2408.13275}
\BIBentrySTDinterwordspacing

\end{thebibliography}

\end{document}